\documentclass[conference]{IEEEtran}
\IEEEoverridecommandlockouts
\usepackage{cite}
\usepackage{amsmath,amssymb,amsfonts}
\usepackage{algorithm} 
\usepackage{algorithmic}
\usepackage{graphicx}
\usepackage{textcomp}
\usepackage{xcolor}
\usepackage{subcaption}
\usepackage{multicol}
\usepackage{wrapfig}
\usepackage{adjustbox}
\usepackage{stackengine}\setstackEOL{\cr} %EOL is abbreviation for "end of line."
\usepackage{listings}
\usepackage{color}
\usepackage{multicol}
\usepackage{url}

\def\BibTeX{{\rm B\kern-.05em{\sc i\kern-.025em b}\kern-.08em
    T\kern-.1667em\lower.7ex\hbox{E}\kern-.125emX}}
\begin{document}

\title{Enabling Continual Learning with \\ Differentiable Hebbian Plasticity
\thanks{Work done while Vithursan Thangarasa was an intern at Uber AI.}
}

\author{\IEEEauthorblockN{Vithursan Thangarasa}
\IEEEauthorblockA{\textit{School of Engineering, University of Guelph} \\
\textit{Vector Institute for Artificial Intelligence}\\
Guelph, Canada \\
vthangar@uoguelph.ca}
\and
\IEEEauthorblockN{Thomas Miconi}
\IEEEauthorblockA{\textit{Uber AI} \\
San Francisco, United States \\
tmiconi@uber.com}
\and
\IEEEauthorblockN{Graham W. Taylor}
\IEEEauthorblockA{\textit{School of Engineering, University of Guelph} \\
\textit{Vector Institute for Artificial Intelligence}\\
Guelph, Canada \\
gwtaylor@uoguelph.ca}
}

\maketitle

\begin{abstract}
    Continual learning is the problem of sequentially learning new tasks or
    knowledge while protecting previously acquired knowledge. However,
    catastrophic forgetting poses a grand challenge for neural networks
    performing such learning process. Thus, neural networks that are deployed in
    the real world often struggle in scenarios where the data distribution is
    non-stationary (concept drift), imbalanced, or not always fully available,
    i.e., rare edge cases. We propose a Differentiable Hebbian Consolidation
    model which is composed of a Differentiable Hebbian Plasticity (DHP) Softmax
    layer that adds a rapid learning plastic component (compressed episodic
    memory) to the fixed (slow changing) parameters of the softmax output layer;
    enabling learned representations to be retained for a longer timescale. We
    demonstrate the flexibility of our method by integrating well-known
    task-specific synaptic consolidation methods to penalize changes in the slow
    weights that are important for each target task. We evaluate our approach on
    the Permuted MNIST, Split MNIST and Vision Datasets Mixture benchmarks, and
    introduce an imbalanced variant of Permuted MNIST --- a dataset that
    combines the challenges of class imbalance and concept drift. Our proposed
    model requires no additional hyperparameters and outperforms comparable
    baselines by reducing forgetting.
\end{abstract}


\section{Introduction}
A key aspect of human intelligence is the \textit{ability to continually adapt
and learn} in dynamic environments, a characteristic which is challenging to
embed into artificial intelligence. Recent advances in machine learning (ML)
have shown tremendous improvements in various problems, by learning to solve one
complex task very well, through extensive training on large datasets with
millions of training examples or more. However, most of the ML models that are
used during deployment in the
% assume that a problem space is stationary, whereas in the
real-world are exposed to non-stationarity where the distributions of acquired
data changes over time. Therefore, after learning is complete, and these models
are further trained with new data, responding to distributional changes,
performance degrades with respect to the original data. This phenomenon known as
\textit{catastrophic forgetting} or \textit{catastrophic
interference}~\cite{mccloskey:catastrophic, French1999CatastrophicFI} presents a
crucial problem for deep neural networks (DNNs) that are tasked with continual
learning~\cite{Ring:1994:CLR:221771}, also called lifelong
learning~\cite{THRUN199525, Thrun1998}.
% As a result, these models suffer when being deployed in environments which
% require the model to assimilate knowledge continually to solve various tasks.
% Some of the real-world applications that typically require this kind of
% learning include perception for autonomous vehicles, recommender systems,
% fraud detection, etc.
In continual learning, the goal is to adapt and learn consecutive tasks without
forgetting how to perform well on previously learned tasks, enabling models that
are scalable and efficient over long timescales.\looseness=-1
% DNNs that are capable of automatically adapting to new information without
% having to be retrained on old and new data each time a new task needs to be
% learned, would enable models that are scalable and efficient over long
% timescales.

In most supervised learning methods, DNN architectures require independent and
identically distributed (iid) samples from a stationary training distribution.
However, for ML systems in real-world applications that require continual
learning, the iid assumption is easily violated when: (1) There is concept drift
in the training data distribution. (2) There are imbalanced class distributions
and concept drift occuring simultaneously. (3) Data representing all scenarios
in which the learner is expected to perform are not initially available. In such
situations, learning systems face the ``stability-plasticity dilemma'' which is
a well-known problem for artificial and biological neural
networks~\cite{Carpenter:1987:MPA:28154.28157,Abraham2005MemoryR}. This presents
a continual learning challenge for an ML system where the model needs to provide
a balance between its plasticity (to integrate new knowledge) and stability (to
preserve existing knowledge).

In biological neural networks, synaptic plasticity has been argued to play an
important role in learning and memory~\cite{HOWLAND2008145, pub.1010013091,
Bailey2015StructuralCO} and two major theories have been proposed to explain a
human's ability to perform continual learning. The first theory is inspired by
synaptic consolidation in the mammalian neocortex~\cite{Benna2016} where a
subset of synapses are rendered less plastic and therefore preserved for a
longer timescale. The general idea for this approach is to consolidate and
preserve synaptic parameters that are considered important for the previously
learned tasks. This is normally achieved through task-specific updates of
synaptic weights in a neural network. The second is the complementary learning
system (CLS) theory~\cite{McClellandMcNaughtonOReilly95, KUMARAN2016512}, which
suggests that humans extract high-level structural information and store it in
different brain areas while retaining episodic memories.

Recent work on differentiable plasticity has shown that neural networks with
``fast weights'' that leverage Hebbian learning
rules~\cite{hebb1949organization} can be trained end-to-end through
backpropagation and stochastic gradient descent (SGD) to optimize the standard
``slow weights'', as well as also the amount of plasticity in each synaptic
connection~\cite{DBLP:journals/corr/Miconi16a, conf/icml/MiconiSC18}. These
works use slow weights to refer to the weights normally used to train vanilla
neural networks, which are updated slowly and are often associated with
long-term memory. The fast weights represent the weights that are superimposed
on the slow weights and change quickly from one time step to the next based on
input representations. These fast weights behave as a form of short-term memory
that enable ``reactivation'' of long-term memory traces in the slow weights.
\cite{conf/icml/MiconiSC18} showed that simple plastic networks with learned
plasticity outperform networks with uniform plasticity on various problems.
Moreover, there have been several approaches proposed recently for overcoming
the catastrophic forgetting problem in fixed-capacity models by dynamically
adjusting the plasticity of each synapse based on its importance for retaining
past memories~\cite{PARISI201954}.

Here, we extend the work on differentiable plasticity to the
\textit{task-incremental} continual learning
setting~\cite{DBLP:journals/corr/abs-1904-07734}, where tasks arrive in a
batch-like fashion, and have clear boundaries. We develop a Differentiable
Hebbian Consolidation model that is capable of adapting quickly to changing
environments as well as consolidating previous knowledge by selectively
adjusting the plasticity of synapses. We modify the traditional softmax layer
and propose to augment the slow weights in the final fully-connected (FC) layer
(softmax output layer) with a set of plastic weights implemented using
Differentiable Hebbian Plasticity (DHP).
% The model's slow weights learn deep representations of data and the fast
% weights implemented with DHP learn to quickly bind (``auto-associate'') the
% class labels to representations.
Furthermore, we demonstrate the flexibility of our model by combining it with
recent task-specific synaptic consolidation based approaches to overcoming
catastrophic forgetting such as elastic weight
consolidation~\cite{Kirkpatrick2017OvercomingCF, pmlr-v80-schwarz18a}, synaptic
intelligence~\cite{pmlr-v70-zenke17a} and memory aware
synapses~\cite{Aljundi_2018_ECCV}.  Our model unifies core concepts from Hebbian
plasticity, synaptic consolidation and CLS theory to enable rapid adaptation to
new unseen data, while consolidating synapses and leveraging compressed episodic
memories in the softmax layer to remember previous knowledge and mitigate
catastrophic forgetting. We test our proposed method on established benchmark
problems including the Permuted MNIST~\cite{Goodfellow:2013:MN:3042817.3043084},
Split MNIST~\cite{pmlr-v70-zenke17a} and Vision Datasets
Mixture~\cite{NIPS2018_7631}. We also introduce the Imbalanced Permuted MNIST
problem and show that plastic networks with task-specific synaptic consolidation
methods outperform networks with uniform plasticity.%~\raggedbottom

\section{Relevant Work}

\textbf{Neural Networks with Non-Uniform Plasticity:}
One of the major theories that have been proposed to explain a human's ability
to learn continually is Hebbian learning~\cite{hebb1949organization}, which
% is arguably the simplest and best-known theory for associative learning. It
suggests that learning and memory are attributed to weight plasticity, that is,
the modification of the strength of existing synapses according to variants of
Hebb's rule~\cite{PAULSEN2000172,Song2000, DBLP:journals/scholarpedia/Oja08}.
% While often described in lay terms as \textit{``cells that fire together, wire
% together''},
It is a form of activity-dependent synaptic plasticity where correlated
activation of pre- and post-synaptic neurons leads to the strengthening of the
connection between the two neurons. According to the Hebbian learning theory,
after learning, the related synaptic strength are enhanced while the degree of
plasticity decreases to protect the learned knowledge~\cite{ZENKE2017166}.

Recent approaches in the meta-learning literature have shown that we can
incorporate fast weights into a neural network to perform one-shot and few-shot
learning~\cite{Munkhdalai2018-uo, DBLP:conf/icml/RaeDDL18}.
\cite{Munkhdalai2018-uo} proposed a model that augments FC layers preceding the
softmax with a matrix of fast weights to bind labels to representations. Here,
the fast weights were implemented with \textit{non-trainable} Hebbian
learning-based associative memory. The Hebbian Softmax
layer~\cite{DBLP:conf/icml/RaeDDL18} can improve learning of rare classes by
interpolating between Hebbian learning and SGD updates on the output layer using
a scheduling scheme.

Differentiable plasticity~\cite{conf/icml/MiconiSC18} uses SGD to optimize the
plasticity of each synaptic connection, in addition to the standard fixed (slow)
weights. Here, each synapse is composed of a slow weight and a plastic (fast)
weight that automatically increases or decreases based on the activity over
time.  Although this approach served to be a powerful new method for training
neural networks, it was mainly demonstrated on recurrent neural networks (RNNs)
for solving pattern memorization tasks and maze exploration with reinforcement
learning. Also, these approaches were only demonstrated on meta-learning
problems and not the continual learning challenge of overcoming catastrophic
forgetting. Our work also augments the slow weights in the FC layer with a set
of plastic (fast) weights, but implements these using DHP. We only update the
parameters of the softmax output layer in order to achieve fast learning and
preserve knowledge over time.%~\raggedbottom

\textbf{Overcoming Catastrophic Forgetting:} This work leverages two strategies
 to overcome the catastrophic forgetting problem: 1) \textit{Task-specific
 Synaptic Consolidation} --- Protecting previously learned knowledge by
 dynamically adjusting the synaptic strengths to consolidate and retain
 memories. 2) \textit{CLS Theory} --- A dual memory system where, the neocortex
 (neural network) gradually learns to extract structured representations from
 the data while, the hippocampus (augmented episodic memory) performs rapid
 learning and individuated storage to memorize new instances or experiences.

There have been several notable works inspired by task-specific synaptic
consolidation for mitigating catastrophic
forgetting~\cite{Kirkpatrick2017OvercomingCF,pmlr-v70-zenke17a,Aljundi_2018_ECCV}
and they are often categorized as regularization strategies in the continual
learning literature~\cite{PARISI201954}. All of these regularization approaches
estimate the importance of each parameter or synapse, $\Omega_{k}$, where least
plastic synapses can retain memories for long timescales and more plastic
synapses are considered less important. The parameter importance and network
parameters $\theta_{k}$ are updated in either an online manner or after learning
task $T_n$. Therefore, when learning  new task $T_{n+1}$, a regularizer is added
to the original loss function $\mathcal{L}^{n}(\theta)$, so that we dynamically
adjust the plasticity~w.r.t $\Omega_{k}$ and prevent any changes to important
parameters of previously learned tasks:
\begin{align}
  \label{eq:loss}
  \tilde{\mathcal{L}}^{n}(\theta) = \mathcal{L}^{n}(\theta) + \underbrace{\lambda \sum_{k}\Omega_{k}(\theta_{k}^{n} - \theta_{k}^{n-1})^{2}}_{\text{regularizer}}
\end{align}
\noindent where $\theta_{k}^{n-1}$ are the learned network parameters after
training on the previous ${n-1}$ tasks and $\lambda$ is a hyperparameter for the
regularizer to control the amount of forgetting.

The main difference in these regularization strategies is on the method used to
compute the importance of each parameter, $\Omega_{k}$. Elastic Weight
Consolidation (EWC)~\cite{Kirkpatrick2017OvercomingCF} used the values given by
the diagonal of an approximated Fisher information matrix for $\Omega_{k}$, and
this was computed offline after training on a task was completed. An online
variant of EWC was proposed by~\cite{pmlr-v80-schwarz18a} to improve EWC's
scalability by ensuring the computational cost of the regularization term does
not grow with the number of tasks. Synaptic Intelligence
(SI)~\cite{pmlr-v70-zenke17a} is an online variant for computing the parameter
importance where, $\Omega_{k}$ is the cumulative change in individual synapses
over the entire training trajectory on a particular task. Memory Aware Synapses
(MAS)~\cite{Aljundi_2018_ECCV} is an online method that measures $\Omega_{k}$ by
the sensitivity of the learned function to a perturbation in the parameters,
instead of measuring the change in parameters to the loss as seen in SI and EWC.
% The authors use the cumulative change in individual synapses on the squared
% L2-norm of the penultimate layer hence decoupling the importance parameters
% from labels and enabling the importance parameters to continue to update even
% in absence of labels.

Our work draws inspiration from CLS theory which is a powerful computational
framework for representing memories with a dual memory system via the neocortex
and hippocampus. There have been numerous approaches based on CLS principles
involving pseudo-rehersal~\cite{Robins95catastrophicforgetting,
doi:10.1080/09540090412331271199, DBLP:journals/corr/abs-1802-03875}, exact or
episodic replay~\cite{NIPS2017_7225, DBLP:journals/pami/LiH18a} and generative
replay~\cite{DBLP:conf/nips/ShinLKK17, DBLP:conf/nips/WuHLWWR18}. However, in
our work, we are primarily interested in neuroplasticity techniques inspired
from CLS theory for alleviating catastrophic forgetting. Earlier work from
\cite{Hinton87usingfast, Gardner-Medwin1989} showed how each synaptic connection
can be composed of a fixed weight where slow learning stores long-term knowledge
and a fast-changing weight for temporary associative memory. This approach
involving slow and fast weights is analogous to properties of CLS theory to
overcome catastrophic forgetting during continual learning.
% Moreover,
% the model reduced catastrophic forgetting by using Hebbian learning and an
% optimization rule that decreased overlap when training the long-term memory.
Recent research in this vein has included replacing soft attention mechanism
with fast weights in RNNs~\cite{NIPS2016_6057}, the Hebbian Softmax
layer~\cite{DBLP:conf/icml/RaeDDL18}, augmenting slow weights in the FC layer
with a fast weights matrix~\cite{Munkhdalai2018-uo}, differentiable
plasticity~\cite{DBLP:journals/corr/Miconi16a, conf/icml/MiconiSC18} and
neuromodulated differentiable plasticity~\cite{miconi2018backpropamine}.
However, all of these methods were focused on rapid learning on simple tasks or
meta-learning over a distribution of tasks or datasets. Furthermore, they did
not examine learning a large number of new tasks while, alleviating catastrophic
forgetting in continual learning.%~\raggedbottom

\section{Continual Learning Formulation}~\label{sec:cl}
In the continual learning setup, we train a neural network model on a sequence
of tasks $T_{1:n_{\max}}$, where $n_{\max}$ is the maximum number of tasks the
model is to learn. Unlike the standard supervised learning setup, continual
learning trains a model on data that is fetched in sequential chunks enumerated
by tasks. Therefore, in a continual learning sequence, the model receives a
sequence of tasks $T_{1:n_{\max}}$ that is to be learned, each with its
associated training data $(\mathcal{X}_n, \mathcal{Y}_n)$, where $\mathcal{X}_n$
is the input data and the corresponding label data denoted by $\mathcal{Y}_n$.
Each task $T_n$ has its own task-specific loss $\mathcal{L}^{n}$, that will be
combined with a regularizer loss term (see Eq.~\ref{eq:loss}) to prevent
catastrophic forgetting. After training is complete, the model will have learned
an approximated mapping $f$ to the the true underlying function $\bar{f}$. The
learned $f$ maps a new input $\mathcal{X}$ to the target outputs
$\mathcal{Y}_{1:n}$ for all $T_{1:n}$ tasks the network has learned so far.
% Also, it is to be noted that the set of classes contained in each task can be
% different from each other, as we have done in the SplitMNIST and Vision Datasets
% Mixture benchmarks.

\section{Differentiable Hebbian Consolidation}~\label{sec:dhpmodel}
In our model, each synaptic connection in the softmax layer has two weights: 1)
The slow weights, $\theta \in \mathbb{R}^{m \times d}$, where $m$ is the number
of units in the final hidden layer and $d$ is the number of outputs of the last
layer. 2) A Hebbian plastic component of the same cardinality as the slow
weights, composed of the plasticity coefficient, $\alpha$, and the Hebbian
trace, $\mathrm{Hebb}$. The $\alpha$ is a scaling parameter for adjusting the
magnitude of the $\mathrm{Hebb}$.  The Hebbian traces accumulate the mean hidden
activations of the final hidden layer $h$ for each target label in the
mini-batch $\{y_{1:B}\}$ of size $B$ which are denoted by $\tilde{h} \in
\mathbb{R}^{1 \times m}$ (refer to Algorithm~\ref{alg:dhpsoftmax}). Given the
pre-synaptic activations of neurons $i$ in $h$, we can formally compute the
post-synaptic activations of neurons $j$ using Eq.~\ref{eq:dp} and obtain the unnormalized
log probabilities (softmax pre-activations) $z$. The softmax function is then
applied on $z$ to obtain the desired predicted probabilities $\hat{y}$ thus,
$\hat{y} = \mathrm{softmax}(z)$. The $\eta$ parameter in Eq.~\ref{eq:hebb_rule}
is a scalar value that dynamically learns how quickly to acquire new experiences
into the plastic component, and thus behaves as the ``learning rate'' for the
plastic connections. The $\eta$ parameter also acts as a decay term for the
$\mathrm{Hebb}$ to prevent instability caused by a positive feedback loop in the
Hebbian traces.
\begin{gather}
\label{eq:dp}
\begin{align}
  z_{j} =  \sum_{i=1}^{m} \ (\underbrace{\theta_{i,j}}_{\text{slow}} +  \underbrace{\alpha_{i,j}\mathrm{Hebb}_{i,j}}_{\text{plastic (fast)}})h_{i} \
\end{align}\\
  \label{eq:hebb_rule}
  \mathrm{Hebb}_{i,j} \leftarrow (1-\eta)\mathrm{Hebb}_{i,j} +  \eta \tilde{h}_{i,j}
\end{gather}
\noindent The network parameters $\alpha_{i,j}$, $\eta$ and $\theta_{i,j}$ are
optimized by gradient descent as the model is trained sequentially on different
tasks in the continual learning setup. In standard neural networks the weight
connection has only fixed (slow) weights, which is equivalent to setting
$\alpha =$ 0 in Eq.~\ref{eq:dp}.%~\raggedbottom
\vspace{-5pt}
\begin{figure}[H]
    \begin{algorithm}[H]
      \caption{Batch update Hebbian traces.}\label{alg:dhpsoftmax}
      \begin{algorithmic}[1]
        \STATE {\bfseries Input:} $h_{1:B}$ (hidden activations of penultimate layer), \\
        \hspace{28pt} $y_{1:B}$ (target
        labels), \\
        \hspace{28pt} $\mathrm{Hebb}$
        (Hebbian trace)
        \STATE {\bfseries Output:} $z_{1:B}$ (softmax pre-activations)
        \FOR{each target label $c \in \{y_{1:B}\}$}
        \STATE $s \gets \sum_{b=1}^{B}[y_{b} = c]$ \hfill{\scriptsize{\emph{/*Count total occurences of  c $\in$ y.*/}}}
        \IF{$s > 0$}
        \STATE $\tilde{h} \gets \frac{1}{s}\sum_{b=1}^{B}h[y_b = c]$\hfill{\scriptsize{\emph{/*Update Hebb for class $c$.*/}}}
        \STATE $\mathrm{Hebb}_{:,c} \gets (1 - \eta)\mathrm{Hebb}_{:,c} \ + \  \eta \tilde{h}$
        \ENDIF
        \ENDFOR
        \STATE $z \gets (\theta + \alpha \mathrm{Hebb})h$ \hfill  {\scriptsize{\emph{/*Compute softmax pre-activations.*/}}}
      \end{algorithmic}
    \end{algorithm}
\end{figure}
\subsection{Hebbian Update Rule}
\begin{figure}[t]
    \centering
    \includegraphics[width=\linewidth]{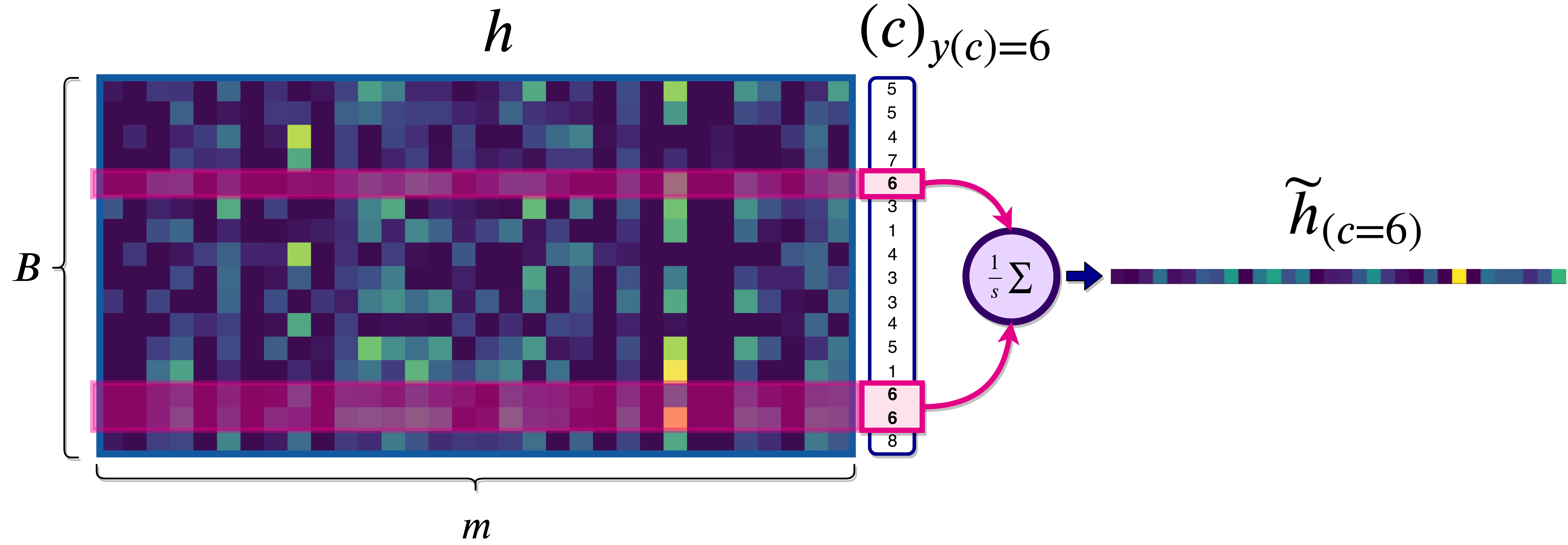}
    \caption{An example of a Hebbian update for the class, $c = $ 6 $\in y_{1:B}$. Here, we are given the hidden activations of the final hidden layer, $h$. Multiple hidden activations corresponding to class $c = $ 6 (represented by the pink boxes) are averaged into one vector denoted by $\tilde{h} \in \mathbb{R}^{1 \times m}$. This Hebbian update visualization reflects Lines 4-6 in Algorithm~\ref{alg:dhpsoftmax} and is repeated for each unique class in the target vector $y_{1:B}$.}
    \label{fig:hebbex}
\end{figure}

The Hebbian traces are initialized to zero only at
the start of learning the first task $T_{1}$ and during training, the
$\mathrm{Hebb}$ is automatically updated in the forward pass using
Algorithm~\ref{alg:dhpsoftmax}. Specifically, the Hebbian update for a
coressponding class $c$ in $y_{1:B}$ is computed on line 6. This Hebbian update
$\frac{1}{s}\sum_{b=1}^{B}h[y_b = c]$ is analogous to another formulaic
description of the Hebbian learning update rule $w_{i,j} =
\frac{1}{N}\sum_{k=1}^{N}a_i^{k}a_j^{k}$~\cite{hebb1949organization}, where
$w_{i, j}$ is the change in weight at connection $i,j$ and $a_{i}^{k}$,
$a_{j}^{k}$ denote the activation levels of neurons $i$ and $j$, respectively,
for the $k$\textsuperscript{th} input. Therefore, in our model, $w = \tilde{h}$
the Hebbian weight update, $a_i = h$ the hidden activations of the last hidden
layer, $a_j = y$ the corresponding target class in $y_{1:B}$ and $N = s$ the
number of inputs for the corresponding class in $y_{1:B}$ (see
Algorithm~\ref{alg:dhpsoftmax} and Figure~\ref{fig:hebbex} for an example). Across the model's lifetime, we only update the
Hebbian traces during training as it learns tasks in a continual manner.
Therefore, during test time, we maintain and use the most recent $\mathrm{Hebb}$
traces to make predictions.

Our model explores an optimization scheme where hidden activations  are
accumulated directly into the softmax output layer weights when a class has been
seen by the network. This results in better initial representations and can also
retain these learned deep representations for a much longer timescale. This is
because memorized activations for one class are not competing for space with
activations from other classes. Fast learning, enabled by a highly plastic
weight component, improves test accuracy for a given task. Between tasks this
plastic component decays to prevent interference, but selective consolidation
into a stable component protects old memories, effectively enabling the model to
\textit{learn to remember} by modelling plasticity over a range of timescales to
form a learned neural memory (see Section~\ref{sec:pmnist}  ablation study). In
comparison to an external memory, the advantage of DHP Softmax is that it is
simple to implement, requiring no additional space or computation. This allows
it to scale easily with increasing number of tasks.

% \begin{wrapfigure}{r}{0.49\linewidth}
%   \vspace{-35pt}
%   \noindent
%   \begin{minipage}{\linewidth}
%   %\begin{figure}[H]
%     \includegraphics[width=\textwidth]{images/five_vision_vert.pdf}
%     \caption{The average test accuracy on a
%     sequence of 5 diffferent vision datasets $T_{n=1:5}$. The average test
%     accuracy over all learned tasks is provided in the legend. The
%     error bars correspond to the SE across 10 trials.}
%     \label{fig:visionmix}
%   %\end{figure}
% \end{minipage}
% \vspace{-10pt}
% \end{wrapfigure}

% \begin{wrapfigure}{r}{\linewidth}
%   %\vspace{-13pt}
%   \begin{minipage}{\linewidth}

The plastic component learns rapidly and performs sparse parameter updates to
quickly store memory traces for each recent experience without interference from
other similar recent experiences. Furthermore, the hidden activations
corresponding to the same class, $c$, are accumulated into one vector
$\tilde{h}$, thus forming a compressed episodic memory in the Hebbian traces to
reflect individual episodic memory traces (similar to the hippocampus in
biological neural networks~\cite{Chadwick2010, Schapiro2017ComplementaryLS}). As
a result, this method improves learning of rare classes and speeds up binding of
class labels to deep representations of the data without introducing any
additional hyperparameters.%~\raggedbottom
% In Appendix~\ref{app:pytorchdhp}, we provide a sample implementation of the
% DHP Softmax using PyTorch.

\subsection{Hebbian Synaptic Consolidation}
Following the existing regularization strategies such as
EWC~\cite{Kirkpatrick2017OvercomingCF}, Online EWC~\cite{pmlr-v80-schwarz18a},
SI~\cite{pmlr-v70-zenke17a} and MAS~\cite{Aljundi_2018_ECCV}, we regularize
the loss $\mathcal{L}(\theta)$ as in Eq.~\ref{eq:loss} and update the synaptic
importance parameters of the network in an online manner. We rewrite
Eq.~\ref{eq:loss} to obtain the updated quadratic loss for Hebbian Synaptic
Consolidation in Eq.~\ref{eq:dploss} and show that the network parameters
$\theta_{i,j}$ are the weights of the connections between pre- and post-synaptic activity, as seen in Eq.~\ref{eq:dp}.
\begin{align}
  \label{eq:dploss}
  \tilde{\mathcal{L}}^{n}(\theta, \alpha, \eta) = \mathcal{L}^{n}(\theta, \alpha, \eta) + \lambda \sum_{i,j}\Omega_{i,j}(\theta_{i,j}^{n} - \theta_{i,j}^{n-1})^{2}
\end{align}
We adapt the existing task-specific consolidation approaches to our model and do
not compute the synaptic importance parameters on the plastic component of the
network, hence we only regularize the slow weights of the network. Furthermore,
when training the first task $T_{n=1}$, the synaptic importance parameter,
$\Omega_{i,j}$ in Eq.~\ref{eq:dploss}, was set to 0 for all of the task-specific
consolidation methods that we tested on except for SI. This is because SI is the
only method we evaluated that estimates $\Omega_{i,j}$ while training, whereas
Online EWC and MAS compute $\Omega_{i,j}$ after learning a task. The plastic
component of the softmax layer in our model can alleviate catastrophic
forgetting of consolidated classes by allowing gradient descent to optimize how
plastic the connections should be (i.e.~less plastic to preserve old information
or more plastic to quickly learn new information).%~\raggedbottom

\section{Experiments}

\begin{figure*}[ht]
  \begin{subfigure}[b]{0.49\textwidth}
    \centering
    \includegraphics[width=0.82\textwidth]{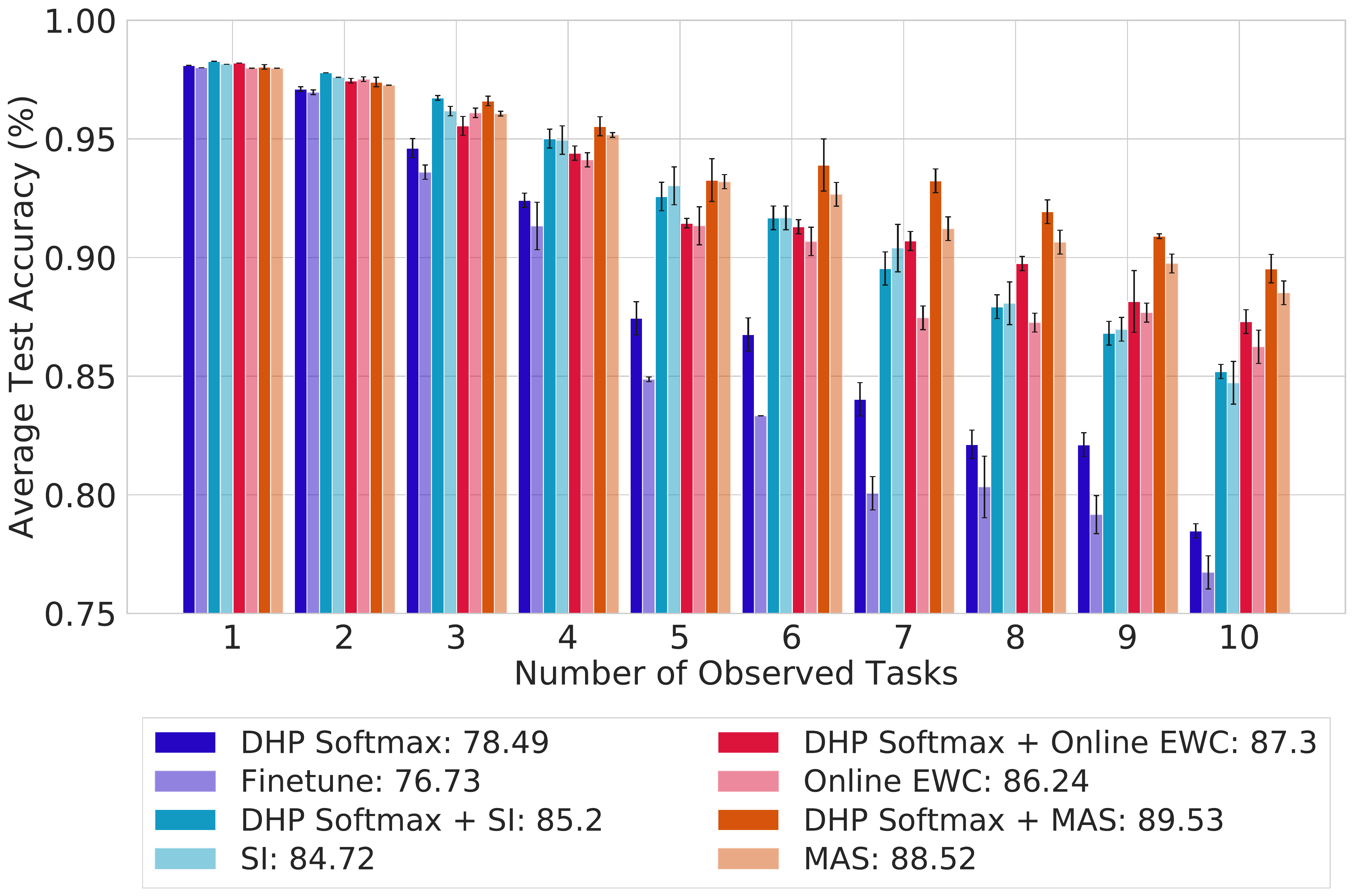}%imb_permuted_mnist_vert
    \caption[]{}\label{fig:pmnist}
  \end{subfigure}
  %\hfill
  \begin{subfigure}[b]{0.49\textwidth}
    \centering
    \raisebox{.00\textwidth}{%
                  \includegraphics[width=0.82\textwidth]{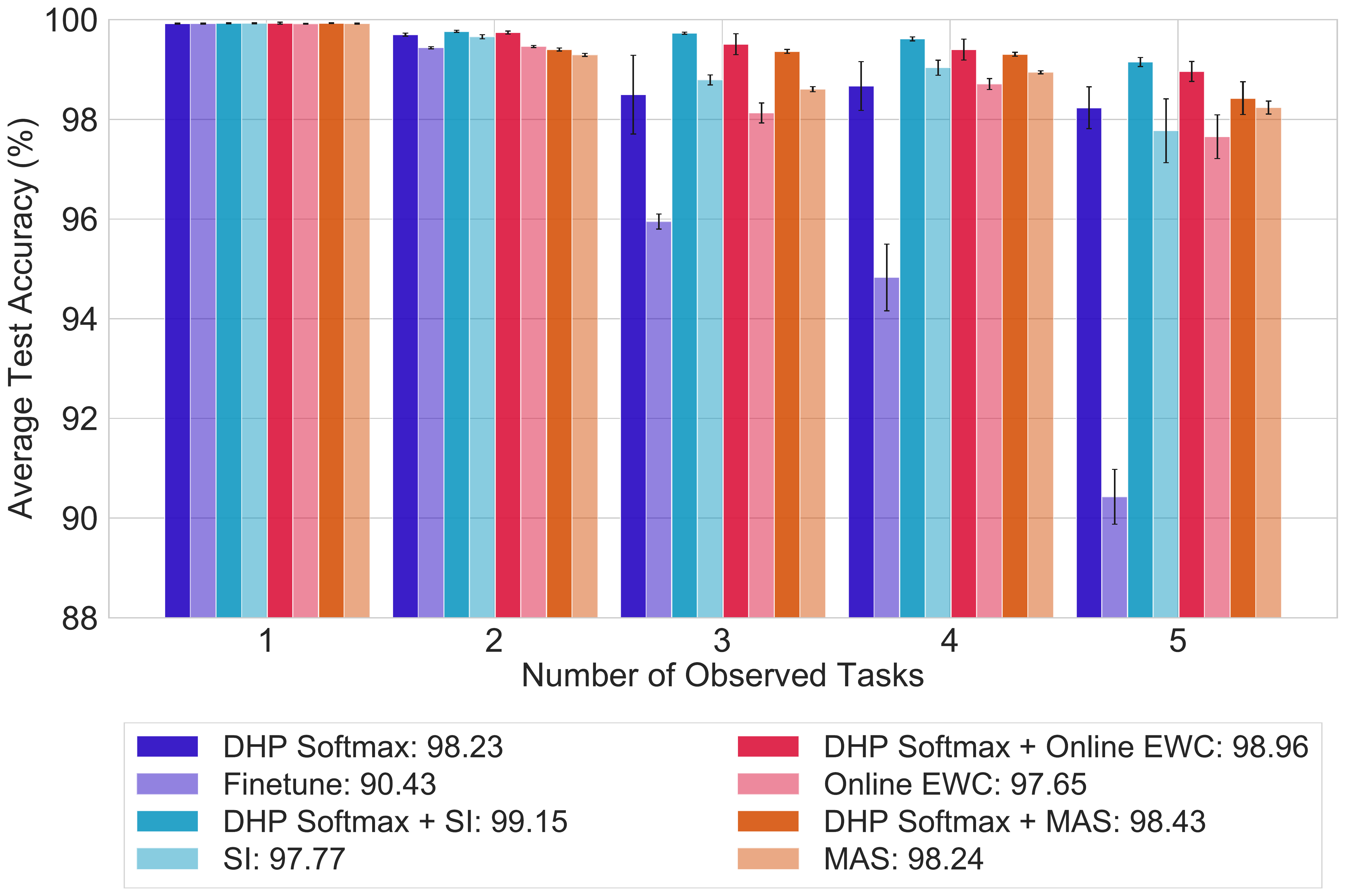}%{images/imb_permuted_mnist_vert.pdf}%
                }
    \caption[]{}\label{fig:splitmnist}
  \end{subfigure}
  \caption[ Global caption ]
  {(a) The average test accuracy on a sequence of $T_{n=1:10}$  Permuted MNIST tasks and (b) $T_{n=1:5}$ binary classification
  tasks from the MNIST dataset. The average test
  accuracy over all learned tasks is provided in the legend. The addition of DHP in all cases improves the model's ability to reduce forgetting. Error bars
  represent the SEM across 10 trials.}
\end{figure*}

In our experiments, we compare our approach to vanilla neural networks with
Online EWC, SI and MAS. Since our approach increases the capacity of the DNN due
to the addition of plastic weights, we add an extra set of slow weights to the
softmax output layer of the standard neural network to match the capacity. We do
this to show that it is not the increased model capacity from the plastic
weights that is helping mitigate the forgetting when performing sequential task
learning, thus ensuring a fair evaluation. We tested our model on the Permuted
MNIST, Split MNIST and Vision Datasets Mixture benchmarks, and also introduce
the Imbalanced Permuted MNIST problem.
% which
% combines the challenges of both class imbalance and concept drift, something
% that a real-world ML system would experience.

For all of the benchmarks, we evaluated the model based on the average
classification accuracy on all previously learned tasks as a function of $n$,
the number of tasks trained so far. To determine memory retention and
flexibility of the model, we are particularly interested in the test performance
on the first task and the most recent one. We also measure forgetting using the
backward transfer metric, BWT $= \frac{1}{T-1}\sum_{i=1}^{T-1} R_{T,i} -
R_{i,i}$~\cite{NIPS2017_7225}, which indicates how much learning new tasks has
influenced the performance on previous tasks. $R_{T,i}$ is the test
classification accuracy on task $i$ after sequentially finishing learning the
$T$\textsuperscript{th} task. While BWT $<$ 0 directly reports catastrophic
forgetting, BWT $>$ 0 indicates that learning new tasks has helped with the
preceding tasks. To establish a baseline for comparison of well-known
task-specific consolidation methods, we trained neural networks with Online EWC,
SI and MAS, respectively, on all tasks in a sequential manner. The
hyperparameters of the consolidation methods (i.e.~EWC, SI and MAS) remain the
same with and without DHP Softmax, and the plastic components are not
regularized. To find the best hyperparameter combination for each of these
synaptic consolidation methods, we performed a grid search using a task sequence
determined by a single seed. Descriptions of the hyperparameters and other
details for all benchmarks can be found in Appendix~\ref{app:hyp}. All
experiments were run on a Nvidia RTX 2080 Ti.%~\raggedbottom

\subsection{Permuted MNIST}\label{sec:pmnist}

\begin{figure*}[ht]
  \begin{subfigure}[b]{0.32\textwidth}
    \centering
    \includegraphics[width=0.80\textwidth]{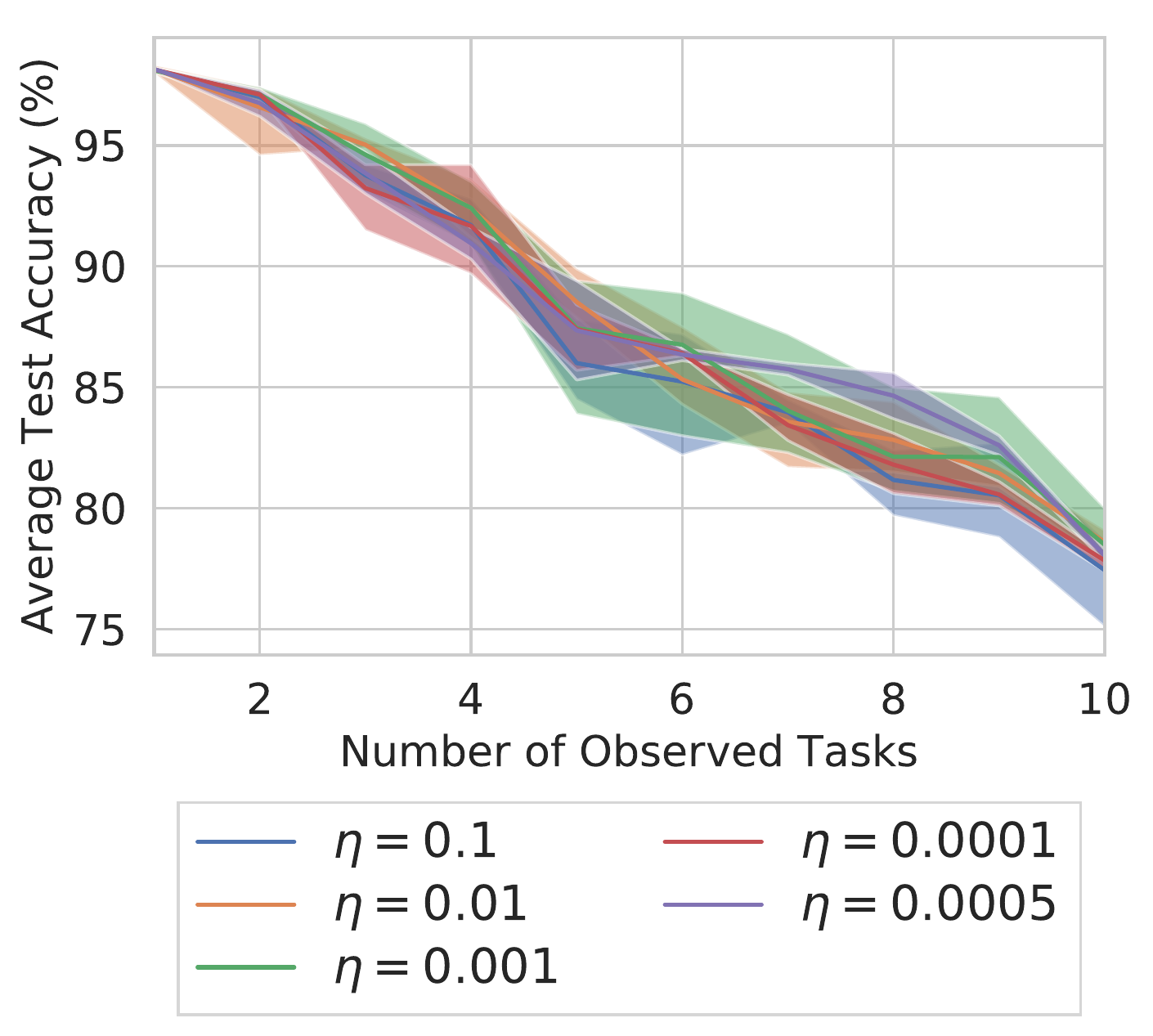}
  \end{subfigure}
  %\hfill
  \begin{subfigure}[t]{0.32\textwidth}
    \centering
    \includegraphics[width=0.80\textwidth]{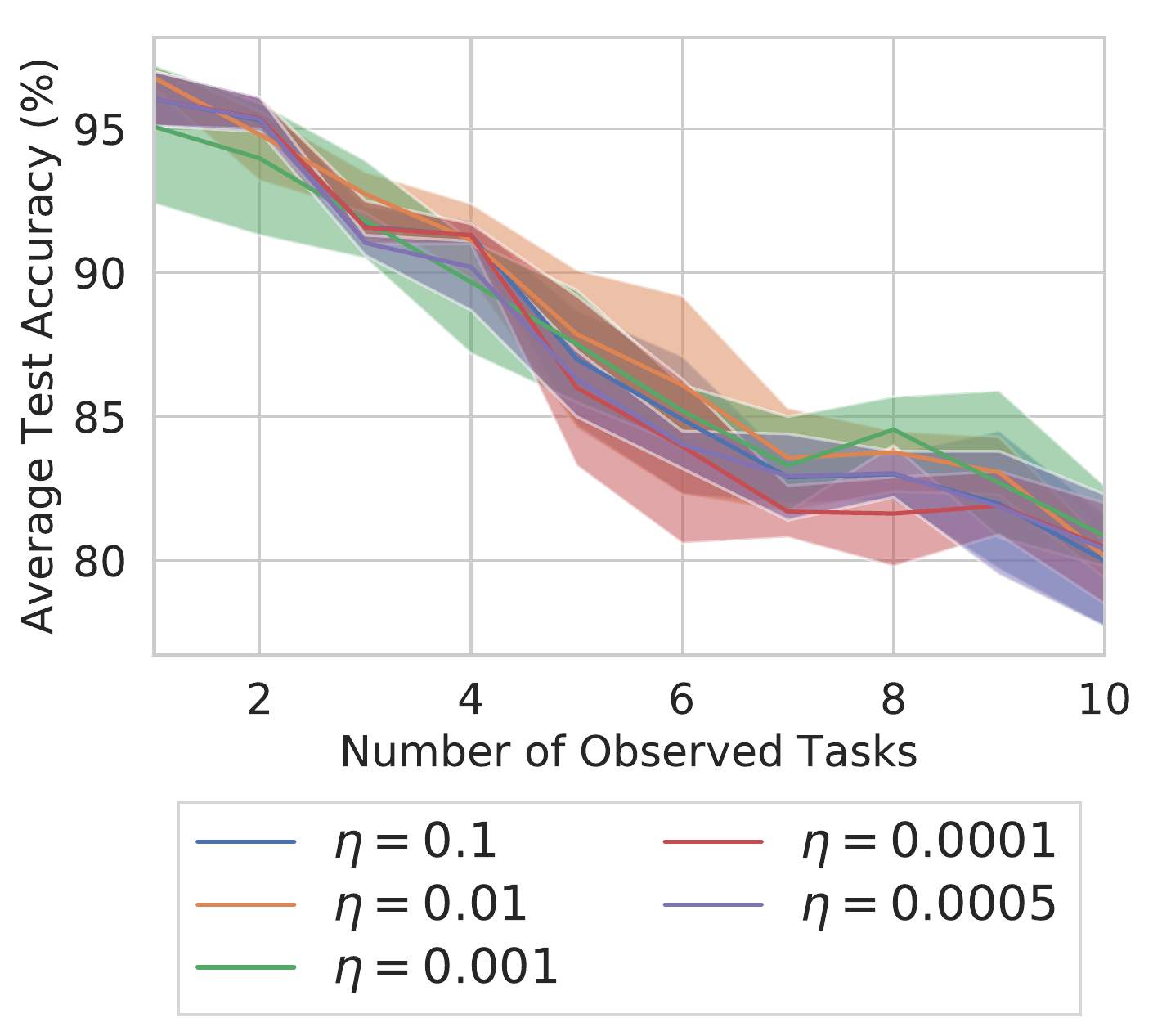}
  \end{subfigure}
  \begin{subfigure}[t]{0.32\textwidth}
    \centering
    \includegraphics[width=0.80\textwidth]{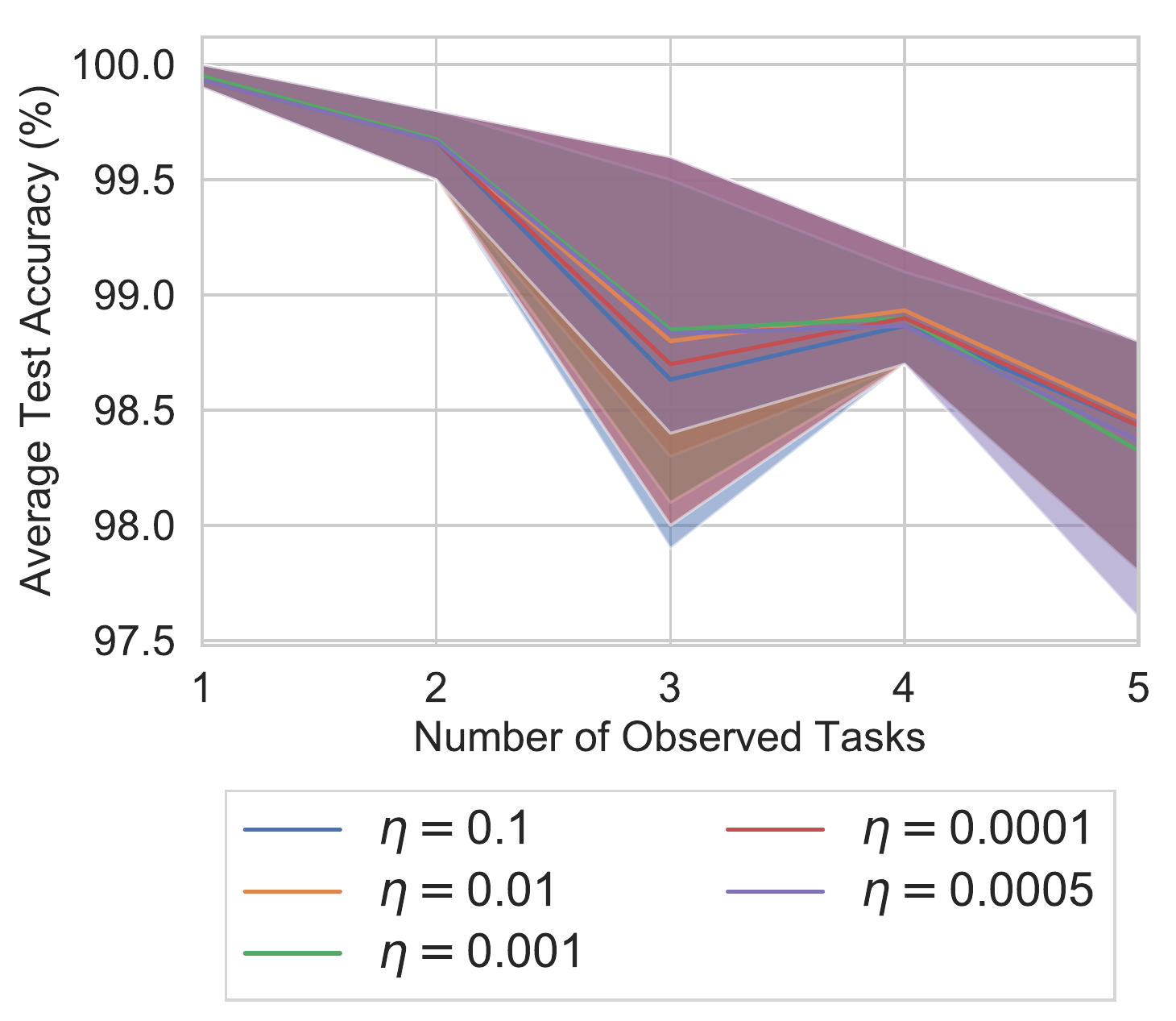}
  \end{subfigure}
  \caption[ Global caption ]
  {A sensitivity analysis on the $\mathrm{Hebb}$ decay term $\eta$ in Eq.~\ref{eq:hebb_rule}. We show the average test accuracy for different initial values of $\eta$ after learning all tasks on the (left) Permuted MNIST, (center) Imbalanced Permuted MNIST and (right) Split MNIST problems. The shaded regions correspond to the standard error of mean (SEM) across 5 trials.}
  \label{fig:senseta}
\end{figure*}

In this benchmark, all of the MNIST pixels are permuted differently for each
task with a fixed random permutation. Although the output domain is constant,
the input distribution changes between tasks and is mostly independent of each
other, thus, there exists a concept drift. In the Permuted MNIST and Imbalanced
Permuted MNIST benchmarks we use a multi-layered perceptron (MLP) network with
two hidden layers consisting of 400 LeakyReLU nonlinearities, and a cross-entropy
loss. The $\eta$ of the plastic component was set to be a value of 0.001 and we
emphasize that we spent little to no effort on tuning the initial value of this
parameter. We swept through a range of values
$\eta \in \{$0.1, 0.01, 0.001, 0.0005, 0.0001$\}$ and found that setting $\eta$
to low values led to the best performance in terms of being able to alleviate
catastrophic forgetting (refer to Figure~\ref{fig:senseta}).

We first compare the performance between our network with DHP Softmax and a
fine-tuned vanilla MLP network we refer to as \textit{Finetune} in
Figure~\ref{fig:pmnist} and no task-specific consolidation methods involved. The
network with DHP Softmax alone showed improvement in its ability to alleviate
catastrophic forgetting across all tasks compared to the baseline network. Then
we compared the performance with and without DHP Softmax using the same
task-specific consolidation methods. Figure~\ref{fig:pmnist} shows the average
test accuracy as new tasks are learned for the best hyperparameter combination
for each task-specific consolidation method. We find our DHP Softmax with
consolidation maintains a higher test accuracy throughout sequential training of
tasks than without DHP Softmax.

\begin{figure}[ht]
  \centering
  \includegraphics[width=\linewidth]{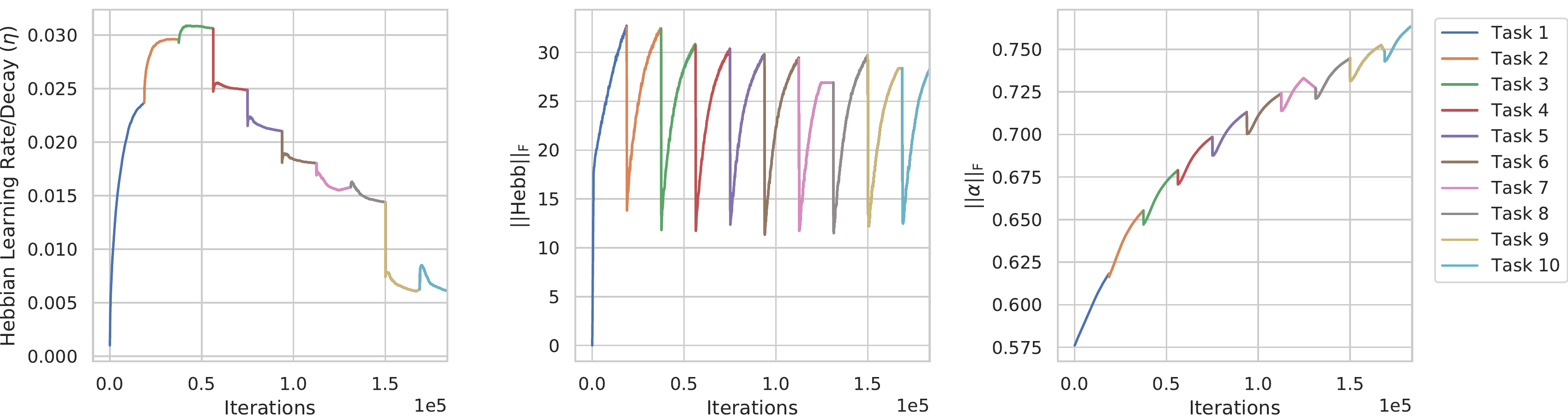}
  \caption{
    \textbf{(left)} Hebbian learning rate and decay value $\eta$, \textbf{(middle)} Frobenius Norm of the Hebbian memory traces $\|\mathrm{Hebb}\|_{\mathrm{F}}$, \textbf{(right)} Frobenius Norm of the plasticity coefficients $\|\alpha\|_{\mathrm{F}}$ while training each task $T_{1:10}$.}
  \label{fig:dhp_graphs}
\end{figure}

% \subsubsection{Ablation Study}
\textbf{Ablation Study:}
We further examine the structural parameters of the network and $\mathrm{Hebb}$
traces to provide further interpretability into the behaviour of our proposed
model. The left plot in Figure~\ref{fig:dhp_graphs} shows the behaviour of
$\eta$ during training as 10 tasks in the Permuted MNIST benchmark are learned
continually. Initially, in task $T_1$, $\eta$ increases very quickly from 0.001
to 0.024 suggesting that the synaptic connections become more plastic to quickly
acquire new information. Eventually, $\eta$ decays after the
3\textsuperscript{rd} task to reduce the degree of plasticity to prevent
interference between the learned representations. We also observe that within
each task from $T_4$ to $T_{10}$, $\eta$ initially increases then decays. The
Frobenius Norm of the $\mathrm{Hebb}$ trace (middle plot in
Figure~\ref{fig:dhp_graphs}) suggests that $\mathrm{Hebb}$ grows
\textit{without} runaway positive feedback every time a new task is learned,
maintaining a memory of which synapses contributed to recent activity. The
Frobenius Norm of $\alpha$ (right plot in Figure~\ref{fig:dhp_graphs}) indicates
that the plasticity coefficients grow within each task, thus the
network continuously leverages the structure in the plastic component. It is important to
note that SGD and backpropagation are used as
\textit{meta-learning} to tune the structural parameters in the plastic
component.%~\raggedbottom

\subsection{Imbalanced Permuted MNIST}
We introduce the Imbalanced Permuted MNIST problem which is identical to the
Permuted MNIST benchmark but, now each task is an imbalanced distribution where
training samples in each class were artificially removed based on some random
probability (see Appendix~\ref{app:hyp}). This benchmark was motivated by
the fact that class imbalance and concept drift can hinder predictive
performance, and the problem becomes particularly challenging when they occur
simultaneously. We see that
DHP Softmax achieves 80.85$\%$ after learning 10 tasks with imbalanced class
distributions in a sequential manner, thus providing significant 4.41$\%$
improvement over the standard neural network baseline of 76.44$\%$. The
significance of the compressed episodic memory mechanism in the Hebbian traces
is more apparent in this benchmark because the plastic component allows rare
classes that are encountered infrequently to be remembered for a longer period
of time.
% Similar to the Permuted MNIST benchmark, we also train neural networks with
% DHP Softmax plus EWC, SI and MAS, then compare the performance against their
% respective baselines.
We find that DHP Softmax with MAS achieves a 0.04 decrease in BWT, resulting in
an average test accuracy of 88.80$\%$ and a 1.48$\%$ improvement over 
MAS alone. %also outperforming all other methods.~\raggedbottom

\subsection{Split MNIST}~\label{sec:splitmnist}
We split the original MNIST
dataset~\cite{lecun-gradientbased-learning-applied-1998} into a sequence of 5
binary classification tasks: $T_{1}=$ \{0/1\}, $T_{2}=$ \{2/3\}, $T_{3}=$
\{4/5\}, $T_{4}=$ \{6/7\} and $T_{5}=$ \{8/9\}. The output spaces are disjoint
between tasks, unlike the previous two benchmarks. Similar to the network used
by~\cite{pmlr-v70-zenke17a}, we use an MLP network with two hidden layers of
256 LeakyReLU nonlinearities each, and a cross-entropy loss. The initial $\eta$ value
was set to 0.001 as seen in previous benchmark experiments. We observed that DHP
Softmax alone achieves 98.23$\%$ thus, provides a 7.80$\%$ improvement on test
performance compared to a finetuned MLP network (Figure~\ref{fig:splitmnist}).
Also, combining DHP Softmax with task-specific consolidation consistently
decreases BWT, leading to a higher average test accuracy across all tasks,
especially the most recent one, $T_{5}$.%~\raggedbottom

% A multi-headed approach
% was used to avoid interference between digits at the softmax output layer due to
% changes in the label distribution. We compute the cross-entropy loss,
% $L(\theta)$, at the softmax output layer for the digits present in the current
% task, $T_n$.

% We train the network on a sequence of $T_{n=1:5}$ tasks with mini-batches of
% size 64 and optimized using plain SGD with a fixed learning rate of 0.01 for 10
% epochs.

\subsection{Vision Datasets Mixture}\label{sec:cvmix}
%\vskip -0.2in
% \begin{wrapfigure}{r}{0.49\linewidth}
%   \vspace{-35pt}
%   \noindent
%   \begin{minipage}{\linewidth}
%   %\begin{figure}[H]
%     \includegraphics[width=\textwidth]{images/five_vision_vert.pdf}
%     \caption{The average test accuracy on a
%     sequence of 5 diffferent vision datasets $T_{n=1:5}$. The average test
%     accuracy over all learned tasks is provided in the legend. The
%     error bars correspond to the SE across 10 trials.}
%     \label{fig:visionmix}
%   %\end{figure}
% \end{minipage}
% \vspace{-10pt}
% \end{wrapfigure}

Following previous works~\cite{NIPS2018_7631, 1803.10123}, we perform continual
learning on a sequence of 5 vision datasets: MNIST, notMNIST\footnote{Originally
published at \url{http://yaroslavvb.blogspot.com/2011/09/notmnist-dataset.html}
and downloaded from \url{https://github.com/davidflanagan/notMNIST-to-MNIST}.},
FashionMNIST~\cite{Xiao2017FashionMNISTAN}, SVHN~\cite{37648} and
CIFAR-10~\cite{Krizhevsky09learningmultiple}.  The MNIST, notMNIST and
FashionMNIST datasets are zero-padded to be of size 32$\times$32 and are
replicated 3 times to create grayscale images with 3 channels, thus matching the
resolution of the SVHN and CIFAR-10 images. 

Here, we use a CNN architecture that is similar to the one used
in~\cite{NIPS2018_7631, 1803.10123}, which consists of 2 convolutional layers
with 20 and 50 channels respectively, and a kernel size of 5. Each convolution
layer is followed by LeakyReLU nonlinearities (negative threshold of 0.3) and
2$\times$2 max-pooling operations with stride 2. The two convolutional layers
are followed by an FC layer of size 500 before the final softmax output layer. The initial $\eta$ parameter value was set to
0.0001. We train the network with mini-batches of size 32 and optimized using
plain SGD with a fixed learning rate of 0.01 for 50 epochs per task.

We found that DHP Softmax plus MAS decreases BWT by 0.04 resulting in a 2.14$\%$
improvement in average test accuracy over MAS on its own (see
Table~\ref{tab:allstats}). Also, SI with DHP Softmax outperforms other
competitive methods with an average test performance of 81.75$\%$ and BWT of
-0.04 after learning all five tasks. In Table~\ref{tab:allstats}, we present a
summary of the final average test performance after learning all tasks in the
respective continual learning problems. Here, we summarize the average test
accuracy and BWT across ten trials for each of the benchmarks.

\begin{table*}[htbp]
  \caption{The average test accuracy ($\%$, higher is better) and backward transfer (BWT, lower is better) after learning all tasks on each benchmark, respectively. The results are averaged over 10 trials.}
  \begin{center}
  \begin{tabular}{|c|c|c|c|c|}
  \hline
  \textbf{Method}& \textbf{Permuted-MNIST}& \textbf{\addstackgap{\stackanchor{Imbalanced}{Permuted-MNIST}}}& \textbf{SplitMNIST} & \textbf{5-Vision Mixture} \\
  \hline
  Finetune & 76.73 / -0.19 & 76.44  / -0.20 & 90.43 / -0.13 & 60.02 / -0.33 \\
  \hline
  DHP Softmax & 78.49 / -0.16 & 80.85 / -0.14 & 98.23 / -0.02 & 62.94 / -0.26 \\
  \hline
  SI & 84.72 / -0.13 & 85.92 / -0.06 & 97.77  / -0.04 & 81.26 / -0.06 \\
  \hline
  DHP Softmax + SI & 85.20  / -0.09 & 85.39 / -0.06 & \textbf{99.15  / 0.00} & \textbf{81.75 / -0.04} \\
  \hline 
  Online EWC & 86.24 / -0.11   & 87.18 / -0.09 & 97.65  / -0.03 & 78.61 / -0.07 \\
  \hline
  DHP Softmax + Online EWC & 87.30  / -0.09 & 87.43 / -0.08             & 98.96  / -0.01 & 79.10  / -0.04 \\
  \hline
  MAS & 88.52  / -0.08  & 87.32   / -0.09              & 98.24  / -0.02 & 78.51  / -0.05  \\
  \hline
  DHP Softmax + MAS        & \textbf{89.53  / -0.06}  &\textbf{88.80  / -0.05}  & 98.43  / -0.01 & 80.66  / -0.01 \\
  \hline
  \end{tabular}
  \label{tab:allstats}
  \end{center}
\end{table*}

\section{Discussion and Conclusion}
We have shown that the problem of catastrophic forgetting in continual learning
environments can be alleviated by adding compressed episodic memory in the
softmax layer through DHP and performing task-specific updates on synaptic
parameters based on their individual importance for solving previously learned
tasks. The compressed episodic memory allows new information to be learned in
individual traces without overlapping representations, thus avoiding
interference when added to the structured knowledge in the slow changing weights
and allowing the model to generalize across experiences. The $\alpha$ parameter
in the plastic component automatically learns to scale the magnitude of the
plastic connections in the Hebbian traces, effectively choosing when to be less
plastic (protect old knowledge) or more plastic (acquire new information
quickly). The neural network with DHP Softmax showed noticeable improvement
across all benchmarks when compared to a neural network with a traditional
softmax layer. The DHP Softmax
does not introduce any additional hyperparameters since all of the structural
parameters of the plastic part $\alpha$ and $\eta$ are learned, and setting the
initial $\eta$ value required very little tuning effort.

We demonstrated the flexibility of our model where, in addition to DHP Softmax,
we can perform Hebbian Synaptic Consolidation by regularizing the slow weights
using EWC, SI or MAS to improve a model's ability to alleviate catastrophic
forgetting after sequentially learning a large number of tasks with limited
model capacity. DHP Softmax combined with SI outperforms other consolidation
methods on the Split MNIST and 5-Vision Datasets Mixture. The approach where we
combine DHP Softmax and MAS consistently leads to overall superior results
compared to other baseline methods on the Permuted MNIST and Imbalanced Permuted
MNIST benchmarks. This is interesting because the local variant of MAS does
compute the synaptic importance parameters of the slow weights $\theta_{i,j}$
layer by layer based on Hebb's rule, and therefore synaptic connections $i,j$
that are highly correlated would be considered more important for the given task
than those connections that have less correlation. Furthermore, our model
consistently exhibits lower negative BWT across all benchmarks, leading to
higher average test accuracy over methods without DHP. This gives a strong
indication that Hebbian plasticity enables neural networks to learn continually
and remember distant memories, thus reducing catastrophic forgetting when
learning from sequential datasets in dynamic environments. Furthermore,
continual synaptic plasticity can play a key role in learning from limited
labelled data while being able to adapt and scale at long timescales. We hope
that our work will open new investigations into gradient descent optimized
Hebbian consolidation for learning and memory in DNNs to enable continual
learning.~\raggedbottom

\bibliographystyle{myIEEEtran}
\bibliography{refs.bib}

% \section{Details on experimental setup and hyperparameter settings}~\label{app:experiments}
\appendix

\subsection{Details on Hyperparameters and Experimental Setup}~\label{app:hyp}
% \subsection{Permuted MNIST Experiments}~\label{app:pmnist}

\noindent \textbf{Permuted MNIST: }
We train the network on a sequence of tasks $T_{n=1:10}$ with
mini-batches of size 64 and optimized using plain SGD with a learning rate of
0.01 for 20 epochs on each task. The regularization hyperparameter $\lambda$ for each of
the task-specific consolidation methods is set to $\lambda = 100$  for Online
EWC~\cite{pmlr-v80-schwarz18a}, $\lambda = 0.1$ for
SI~\cite{pmlr-v70-zenke17a} and $\lambda = 0.1$ for
MAS~\cite{Aljundi_2018_ECCV}. We note that for the SI method, $\lambda$ refers
to the parameter $c$ in the original work~\cite{pmlr-v70-zenke17a} but we use
$\lambda$ to keep the notation consistent across other task-specific
consolidation methods.  In SI, the damping parameter, $\xi$, was set to 0.1.\\

%  To
% find the best hyperparameter combination for each of these synaptic
% consolidation methods, we performed a grid search using a task sequence
% determined by a single seed. For Online EWC, we tested values of $\lambda \in
% \{$10, 20, 50,\ldots, 400$\}$, SI --- $\lambda \in \{$0.01, 0.05,\ldots, 0.5,
% 1.0$\}$ and MAS --- $\lambda \in \{$0.01, 0.5, \ldots, 1.5, 2.0$\}$.
%~\label{app:imbpmnist}
\noindent \textbf{Imbalanced Permuted MNIST: }
For each task, we
artificially removed training samples from each class in the original MNIST
dataset~\cite{lecun-gradientbased-learning-applied-1998} based on some random
probability. For each class and each task, we draw a different removal
probability from a standard uniform distribution $U(0,1)$, and then remove each
sample from that class with that probability. The distribution of classes in
each dataset corresponding to tasks $T_{n=1:10}$. The $\lambda$ for each
of the task-specific consolidation methods is $\lambda = 400$  for Online
EWC~\cite{pmlr-v80-schwarz18a}, $\lambda = 1.0$ for
SI~\cite{pmlr-v70-zenke17a} and $\lambda = 0.1$ for
MAS~\cite{Aljundi_2018_ECCV}. In SI, the damping parameter, $\xi$, was set to
0.1. Across all experiments, we maintained the same
random probabilities detemined by a single seed to artificially remove training
samples from each class.\\
\noindent \textbf{Split MNIST: }
For the Split MNIST experiments shown in
Figure~\ref{fig:splitmnist}, the regularization hyperparameter $\lambda$ for
each of the task-specific consolidation methods is $\lambda = 400$  for Online
EWC~\cite{pmlr-v80-schwarz18a}, $\lambda = 1.0$ for
SI~\cite{pmlr-v70-zenke17a} and $\lambda = 1.5$ for
MAS~\cite{Aljundi_2018_ECCV}. In SI, the damping parameter, $\xi$, was set to
0.001. We train the
network on a sequence of $T_{n=1:5}$ tasks with mini-batches of size 64 and
optimized using plain SGD with a fixed learning rate of 0.01 for 10 epochs for each task. \\

% To find the best hyperparameter combination for each of these synaptic
% consolidation methods, we performed a grid search using the 5 task binary
% classification sequence (0/1, 2/3, 4/5, 6/7, 8/9). For Online EWC, we tested
% values of $\lambda \in \{$1, 25, 50, 100, \ldots,1$\times$10$^{3}$,
% 2$\times$10$^{3}\}$, SI --- $\lambda \in \{$0.1, 0.5, 1.0, \ldots, 5.0$\}$ and
% MAS --- $\lambda \in \{$0.01, 0.05, 1.0,\ldots, 4.5, 5.0$\}$. 

% \subsubsection*{Hyperparameters}\label{app:5vision}
\noindent \textbf{Vision Datasets Mixture: }
For the 5-Vision Datasets Mixture experiments, the regularization hyperparameter $\lambda$ for each
of the task-specific consolidation methods is $\lambda =$ 100 for Online
EWC~\cite{pmlr-v80-schwarz18a}, $\lambda = 0.1$ for
SI~\cite{pmlr-v70-zenke17a} and $\lambda = 1.0$ for
MAS~\cite{Aljundi_2018_ECCV}. In SI, the damping parameter, $\xi$, was set to
0.1.

\end{document}